%% file: main.tex
\documentclass[runningheads]{llncs}

\usepackage{eccv}

\usepackage{eccvabbrv}

\usepackage{graphicx}
\usepackage{booktabs}

\usepackage[accsupp]{axessibility}  

\usepackage{hyperref}

\usepackage{orcidlink}

\usepackage{booktabs}
\usepackage{multirow}
\usepackage{adjustbox}
\usepackage{tabularx}
\usepackage{threeparttable}
\usepackage{xspace}
\usepackage{siunitx}
\usepackage{ulem}
\usepackage[table]{xcolor}
\definecolor{ourgray}{gray}{0.92}

\newcommand{\mjepa}{MJEPA\xspace}

\begin{document}

\title{MJEPA: A Simple and Scalable Joint-Embedding Predictive Architecture for Audio-Visual Learning} 

\titlerunning{MJEPA}

\author{Revant Teotia\inst{1,2} \and
Adrien Bardes\inst{1} \and
Michael Rabbat\inst{1} \and 
Sumit Chopra\inst{2} \and \\
Matthew Muckley\inst{1}\thanks{Equal supervision.} \and 
Nicolas Ballas\inst{1}$^\star$
}

\authorrunning{R.~Teotia et al.}

\institute{$^1$FAIR at Meta \quad $^2$New York University \\
\email{\{rt2741@nyu.edu\}}
}

\maketitle

\input{sec/0_abstract}
\input{sec/1_intro}
\input{sec/2_relatedwork}
\input{sec/3_preliminaries}
\input{sec/4_method}

\input{sec/5_main_results}
\input{sec/6_conclusion}


%
%
\bibliographystyle{splncs04}
\bibliography{main}

\newpage
\input{sec/7_appendix}

\end{document}

%% file: sec/0_abstract.tex
\begin{abstract}
{\tolerance=200 \emergencystretch=1em
    Self-supervised learning from large-scale video data has emerged as a dominant paradigm for visual representation learning. 
    Since audio and visual streams naturally co-occur in video data, extending this success to jointly learn from both modalities is a natural next step, yet it remains challenging. 
    Existing audio-visual self-supervised methods rely on modality-specific encoders and complex combinations of contrastive or reconstruction objectives, limiting cross-modal synergy and scalability. 
    Joint Embedding Predictive Architectures (JEPAs) offer a simple, modality-agnostic alternative, but have to date been applied primarily to individual modalities. 
    We introduce \mjepa, a joint-embedding predictive architecture for audio-visual learning that uses a single, unified encoder for both modalities. 
    Our approach uses only a single predictive objective, applied both within and across modalities. 
    We show that cross-modal prediction is critical: without it, a shared encoder degrades below unimodal baselines; with it, each modality's representation benefits from the other. 
    Our frozen ViT-g model outperforms the best prior frozen baseline by over 6.8 mAP on AudioSet-20K, surpasses fully finetuned models on ESC-50 and FSD50K, and is competitive on video benchmarks despite using 10x less video data.
\par}
\keywords{Self-Supervised Learning \and Representation Learning \and Multi-modal Learning}
\end{abstract}

%% file: sec/1_intro.tex
\section{Introduction}
\label{sec:intro}

\begin{figure*}[!ht]
\centering
\resizebox{\textwidth}{!}{%
\begin{tabular}{@{}c@{\hspace{4pt}}c@{\hspace{4pt}}c@{}}
{\small\textbf{(a) Prior AV-SSL Methods}} & {\small\textbf{(b) MJEPA (Ours)}} & {\small\textbf{(c) Frozen Evaluation Results}} \\[2pt]
\includegraphics[height=5.5cm]{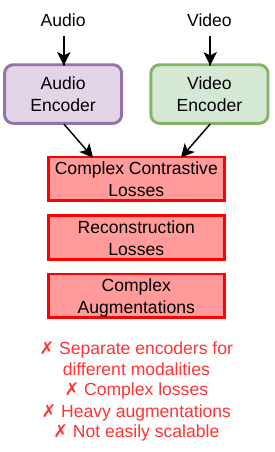} &
\includegraphics[height=5.5cm]{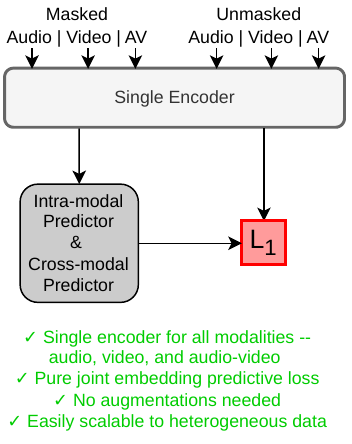} &
\includegraphics[height=5.5cm]{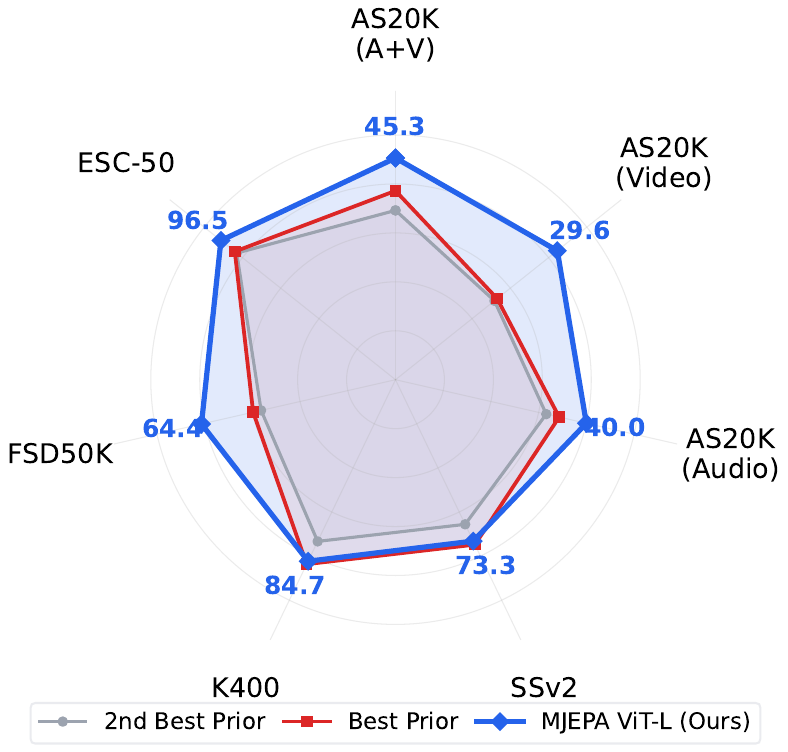}
\end{tabular}%
}
\vspace{-2mm}
\caption{\textbf{Overview of MJEPA.} (a)~Prior AV-SSL methods typically combine modality-specific encoders with a mixture of contrastive losses, reconstruction objectives, and heavy augmentations. (b)~MJEPA adopts a deliberately simple design: a single shared encoder processes all modalities (audio, video, and audio-video) and is trained with only joint-embedding predictive (JEPA) objective. An intra-modal predictor learns modality-specific features and a cross-modal predictor aligns representations across modalities. This design scales naturally to heterogeneous data, including mixtures of unimodal and multimodal sources. (c)~Despite its simplicity, MJEPA produces highly generalizable frozen representations from a single model that match or surpass the best prior frozen features across seven audio, video, and audio-visual benchmarks---outperforming both dedicated unimodal models and specialized multimodal ones.
}
\vspace{-3mm}
\label{fig:teaser}
\end{figure*}

Learning visual representations from large-scale video data has emerged as a dominant paradigm in recent years, with self-supervised methods demonstrating remarkable success across diverse downstream tasks~\cite{vjepa2_2025}.
Videos, however, are inherently multimodal: the visual stream is naturally accompanied by a rich, complementary audio signal that captures information that is often absent from pixels alone.
This observation motivates a natural generalization—extending successful video representation learning to jointly leverage both audio and visual modalities through their intrinsic correspondence in unlabeled video~\cite{look_listen_learn_2017_ICCV}.

Approaches exploring  audio-visual self-supervised learning (AV-SSL) tend to rely on complex, multi-component architectures: separate encoders for each modality combined with intricate loss functions that blend contrastive objectives~\cite{equiav_icml24, cavmae_sync_cvpr25}, masked reconstruction~\cite{georgescu2023avmae}, or hybrids~\cite{cavmae_iclr23, mavil_neurips23, xkd_2022}.
This architectural separation and design complexity potentially limit cross-modal synergy and impede scaling—both in model size and in the use of large-scale unimodal data.

Joint-Embedding Predictive Architectures (JEPAs)~\cite{vjepa_2024, ajepa_2024} offer an alternative that is  explicitly designed to be modality-agnostic: a unified framework where the same objective---predicting representations of masked regions in a learned embedding space---applies seamlessly across input types~\cite{lecun2022path,assran2023self}.
Yet in practice, most existing JEPA methods have only been instantiated for individual modalities (\eg, images, video, audio) leaving the multimodal setting unexplored.

We introduce \textbf{\mjepa} (Multimodal Joint-Embedding Predictive Architecture), a multimodal audio-visual self supervised learning framework that extends the JEPA paradigm to jointly learn from audio and video (Fig.~\ref{fig:teaser}).
Our approach is built on two key design principles: (1) a \textbf{single, unified encoder} that processes audio, video, or paired audio-video inputs within a shared representation space, and (2) a \textbf{single JEPA objective} that can be applied within and across modalities.
Central to our method is the concept of \emph{cross-modal prediction}, which enables positive transfer across modalities.
Beyond standard within-modality JEPA prediction (predicting masked regions from visible context within the same modality), we introduce cross-modal prediction: predicting the representation of one modality from the other.
This objective enforces strong semantic alignment between audio and video representations. Indeed, we show that naively sharing an encoder across modalities \emph{without} cross-modal alignment actually degrades both modalities below their unimodal baselines.
Cross-modal prediction, however, enables positive transfer across modalities.
Combined with scaling to a 1B-parameter ViT-g and heterogeneous data mixtures, we demonstrate that this simple multimodal training learns strong audio-visual representations.

We evaluate \mjepa across audio, video, and audio-visual benchmarks to verify two core hypotheses: (1) a simple, single-encoder SSL approach learns strong audio-visual representation; (2) cross-modal prediction enables positive transfer, where each modality benefits from the other. The resulting frozen representations are competitive with or superior to task-specific finetuned models.
On AudioSet-20K~\cite{audioset_2017}, our frozen ViT-g model outperforms the best prior frozen baseline by over 6.8 mAP on audio-video classification.
On audio benchmarks, our frozen features surpass fully finetuned models on ESC-50~\cite{ESC50_2015} and FSD50K~\cite{fsd50k_2022}.
On video benchmarks Kinetics-400~\cite{kinetics_2017} and SSv2~\cite{ssv2_2017}, incorporating audio data allows our model to nearly match state-of-the-art video-based SSL despite using approximately 10$\times$ less video training data.
\vspace{1em}

In summary, this paper makes the following contributions:
\begin{enumerate}
    \item We introduce \mjepa, a simple and scalable AV-SSL framework that trains a single, unified audio-video encoder with only JEPA objectives --- without reconstruction, contrastive losses, complex augmentations, or supervision.

    \item We demonstrate that cross-modal prediction enables positive transfer across modalities: without it, a shared encoder underperforms unimodal baselines, while with it, each modality's representation improves the other. This synergy allows effective scaling to 1B-parameter models.

    \item Extensive evaluation shows that frozen \mjepa features consistently and substantially outperform prior SSL baselines and are competitive with or superior to fully finetuned state-of-the-art models on several key benchmarks.

    \item Progressive ablations show the contribution of each design component; joint encoding, shared encoder, cross-modal prediction, and scaling; demonstrating that each is essential for learning strong multimodal representations.
\end{enumerate}

%% file: sec/2_relatedwork.tex
\section{Related Work}
\label{sec:related_work}

\paragraph{\textbf{Self-Supervised Learning for Audio and Video.}}
Self-supervised learning has become the dominant paradigm for learning general-purpose representations from unimodal data. 
In vision, masked image modeling~\cite{he2021masked,tong2022videomae}, joint-embedding methods~\cite{oquab2024dinov2,simeoni2025dinov3}, and joint-embedding predictive architectures (JEPAs)~\cite{vjepa_2024,vjepa2_2025} have each produced strong representations, with the latter learning by predicting masked features in a latent space without pixel-level reconstruction. 
In audio, a parallel trajectory has unfolded: the Audio Spectrogram Transformer~\cite{AST_interspeech2021} adapted ViTs to spectrograms, followed by self-supervised methods including data2vec~\cite{baevski2022data2vec}, BEATs~\cite{chen2022beats}, Audio-MAE~\cite{huang2022audiomae}, and A-JEPA~\cite{ajepa_2024}. 
While these unimodal methods learn strong representations, they are trained in isolation and do not leverage the natural correspondence between audio and visual streams.

\paragraph{\textbf{Audio-Visual Self-Supervised Learning.}}
To exploit the co-occurrence of audio and video, a variety of AV-SSL methods have been proposed~\cite{look_listen_learn_2017_ICCV}, typically relying on separate, modality-specific encoders trained with contrastive losses, masked reconstruction, cross-modal distillation, or combinations thereof~\cite{cavmae_iclr23,georgescu2023avmae,mavil_neurips23,cavmae_sync_cvpr25,xkd_2022,equiav_icml24, NEURIPS2021_VATT}. 
EquiAV~\cite{equiav_icml24} relies heavily on data augmentations, using equivariance to encode augmentation-related information into the learned representations. 
The challenge of moving to a shared encoder has been noted: VATT~\cite{NEURIPS2021_VATT}, XKD~\cite{xkd_2022} and AVSiam~\cite{lin2024avsiam} all observe that unimodal representations degrade when replacing separate encoders with a shared one. 
In contrast, \mjepa uses a single shared encoder with only joint embedding predictive objective—requiring no data augmentations, negative samples or reconstruction—and produces representations that are strong both unimodally and multimodally, generalizing to diverse out-of-distribution benchmarks without any fine-tuning.

\paragraph{\textbf{Joint-Embedding Predictive Architectures and Representation Alignment.}}
Our work directly extends the JEPA paradigm~\cite{lecun2022path} to the multimodal domain. 
In JEPA, models learn abstract representations by predicting masked features in a latent space, a principle that has been successfully applied to images (I-JEPA~\cite{assran2023self}), video (V-JEPA~\cite{vjepa_2024}), and audio (A-JEPA~\cite{ajepa_2024}). 
We provide a detailed formulation of the JEPA objective in Sec.~\ref{sec:preliminaries}. 
However, most of these these prior instantiations have remained unimodal. 
\mjepa is the first to propose a simple, unified JEPA with a single shared encoder for audio-visual learning. 
The motivation for a unified model is supported by the Platonic Representation Hypothesis~\cite{platonic_2024}, which posits that representations trained on different modalities converge as models scale. 
Our cross-modal prediction objective explicitly enforces this alignment, a concept shown to be effective in generative modeling~\cite{repa_2024}. 
Other approaches align multiple modalities into a shared space using semantic supervision from image-text or language data~\cite{girdhar2023imagebind,zhu2024languagebind}, whereas \mjepa is purely self-supervised. 
Recent work~\cite{synergy_2025} further shows that one modality can synergize the training of another, providing empirical support for the positive transfer we observe.

%% file: sec/3_preliminaries.tex
\section{Preliminaries}
\label{sec:preliminaries}
Our methodology in Sec.~\ref{sec:method} is presented as a progressive ablation study where we incrementally build our model and evaluate each component. To support this structure, we first introduce the core concepts and experimental setup that are used throughout the subsequent sections.

\subsection{Joint-Embedding Predictive Architectures}
Unlike generative methods that reconstruct raw inputs (\eg, pixels), Joint-Embedding Predictive Architectures (JEPAs)~\cite{lecun2022path} learn abstract representations by performing a prediction task entirely within a learned embedding space. 
The core idea is to predict the representation of a target (a complete view of an input) using only the representation of a context (a partial or corrupted view of the same input). 
This encourages the model to learn semantic features rather than focusing on low-level details.

The typical JEPA setup involves two main components: a context encoder, $E_\theta$, which computes representations for the corrupted input, and a predictor, $P_\phi$, which attempts to predict the target representations from the context representations. 
A key challenge is preventing "representation collapse", a trivial solution where the encoder outputs a constant value. 
This is commonly addressed using an asymmetric training setup, such as a stop-gradient (sg) operation on the target branch and using an exponential moving average (EMA) of the encoder weights to produce stable targets~\cite{grill2020byol, assran2023self}.

Our work builds on the Video JEPA (V-JEPA) model~\cite{vjepa_2024, vjepa2_2025}, which applies this paradigm to video by masking out large blocks of spatio-temporal tubelets. 
The context encoder $E_\theta$ processes the visible tubelets from a masked view $x$, while a target encoder $E_{\bar{\theta}}$ (an EMA of $E_\theta$) processes the complete, unmasked view $y$. 
The predictor $P_\phi$ then takes the output of the context encoder and a set of learnable mask tokens $\Delta_y$ (which specify the positions of the missing patches) to predict the target representations. 
The training objective minimizes the L1 distance between the predicted and target representations for the masked tokens:
\begin{equation}
    \mathcal{L}_{\text{predict}} = \frac{1}{|M|} \sum_{i \in M} \| P_\phi(E_\theta(x), \Delta_y)_i - \text{sg}(E_{\bar{\theta}}(y))_i \|_1,
\end{equation}
where $M$ is a set containing the masked patch indexes from the $x$ view. 
The loss uses a stop-gradient operator, sg, to prevent representation collapse~\cite{grill2020byol} and an exponential moving average $\bar{\theta}$ of $\theta$ is used to update the weight of the target encoder $E_{\bar{\theta}}(y)$ processing the video $y$. 
$\mathcal{L}_{\text{predict}}$ is only applied on masked tokens, and not on the context tokens. 
Our goal is to extend this JEPA paradigm to jointly learn from audio and video within a single, unified architecture.

\subsection{Frozen Evaluation Protocol}
\label{sec:frozen_eval}
To assess the quality of learned representations, we follow the frozen evaluation protocol of~\cite{vjepa_2024, vjepa2_2025}.
A lightweight attentive probe is trained on top of frozen encoder features for each downstream task.
Compared to full fine-tuning, frozen evaluation provides a clearer measure of the intrinsic quality of the learned representations without task-specific adaptation, making it a more reliable assessment of self-supervised learning performance~\cite{sslam_2025}.
Moreover, pretrained encoders are often deployed in a frozen state in practice, making this protocol a more realistic measure of a model's general utility across diverse tasks.
Architectural details of the probe are provided in the supplementary material.

\subsection{Experimental Setting}
We conduct our progressive method development on the AudioSet-20K (AS20K) benchmark~\cite{audioset_2017}, a class-balanced 22k/19k train/eval split of AudioSet~\cite{audioset_2017}. Our base model is a ViT-L~\cite{dosovitskiy2021image} trained on AudioSet2M (AS2M)~\cite{audioset_2017}, which contains about 1.8 million 10-second video clips. 
For \textit{data scaling} experiments, we additionally use the video-only VideoMix2M (VM2M) dataset~\cite{vjepa_2024}, comprising approximately 2 million videos from HowTo100M~\cite{howto100m_2019}, Kinetics-710~\cite{kinetics_2017}, and Something-Something-v2~\cite{ssv2_2017}, and scale our architecture to ViT-g. 
Further details on datasets and training are provided in the supplementary material.

%% file: sec/4_method.tex
\section{MJEPA: Methodology}
\label{sec:method}

\begin{figure*}[t]
    \centering
    \includegraphics[width=\textwidth]{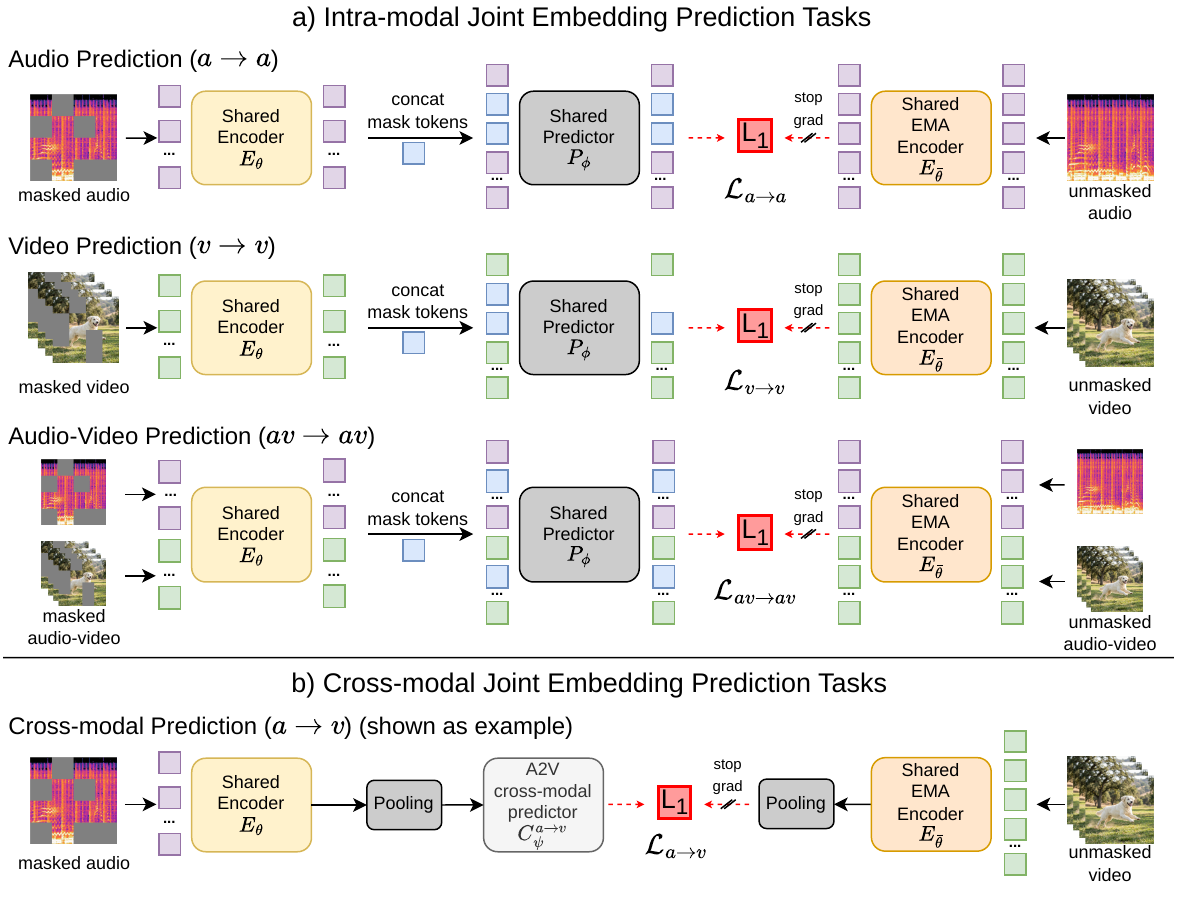}
    \vspace{-9mm}
    \caption{\textbf{\mjepa architecture.} A single shared encoder $E_\theta$ is trained across multiple prediction tasks. \textbf{(a)~Intra-modal prediction:} the encoder and a shared predictor $P_\phi$ are applied independently to three input modes—audio ($a{\to}a$), video ($v{\to}v$), and audio-video ($av{\to}av$)—each predicting masked token representations against targets from an EMA encoder $E_{\bar{\theta}}$. \textbf{(b)~Cross-modal prediction:} shown for $a{\to}v$: pooled last-layer encoder features from the masked source modality are passed through a lightweight cross-modal predictor $C_\psi^{a \to v}$ to predict pooled target features of a different modality. Six such objectives ($a \leftrightarrow v$, $a \leftrightarrow av$, $v \leftrightarrow av$) enforce semantic alignment across modalities. All tasks share the same encoder and EMA encoder weights.}
\vspace{-2mm}
\label{fig:architecture}
\end{figure*}

\begin{figure*}[t]
\centering
\includegraphics[width=\linewidth]{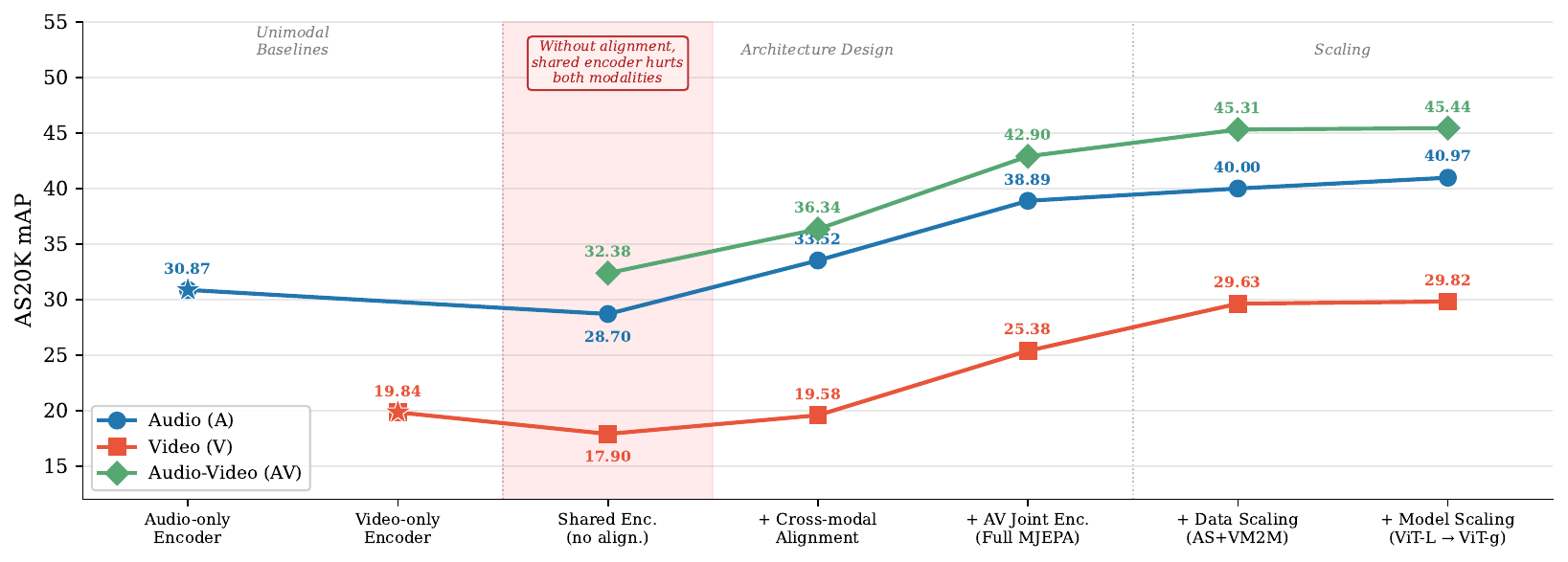}
\vspace{-5mm}
\caption{\textbf{Progressive ablation of architecture and scaling on AS20K.} Starting from unimodal baselines, we first incrementally add components of our architecture (left), then show the effects of data and model scaling (right). A shared encoder without cross-modal alignment (shaded) degrades both modalities below their unimodal baselines. Cross-modal alignment, joint audio-video encoding, and scaling are each critical for learning high-quality, generalizable multimodal representations.}
\vspace{-3mm}
\label{fig:ablation_full}
\end{figure*}

We develop our \mjepa framework incrementally to demonstrate a key finding: while a shared encoder is a natural architecture for multimodal learning, it is not effective without an explicit cross-modal alignment objective. 
We begin by establishing unimodal baselines (Sec.~\ref{sec:unimodal_baselines}), then demonstrate the pitfalls of a naively shared encoder (Sec.~\ref{sec:shared_encoder}). 
We then show how adding our cross-modal alignment objective resolves this issue (Sec.~\ref{sec:cross_modal_alignment}), before introducing the full \mjepa model with joint audio-video encoding (Sec.~\ref{sec:full_mjepa}). 
Finally, we demonstrate the scalability of our approach with respect to both data and model size (Sec.~\ref{sec:scaling}).
Throughout this development, we report audio-only (A), video-only (V), and audio-video (A-V) mean Average Precision (mAP) on AudioSet-20K~\cite{audioset_2017}, a multi-label audio event classification benchmark, using the frozen evaluation protocol from Sec.~\ref{sec:frozen_eval}. 
For unimodal baselines, the encoders receive only the corresponding modality's tokens. For the shared encoder variants, the same model produces features for all three settings by receiving audio, video, or concatenated audio-video tokens as input.
The progressive results are shown in Fig.~\ref{fig:ablation_full}.

\subsection{Unimodal Baselines}
\label{sec:unimodal_baselines}

We first establish unimodal baselines by training two separate, specialized encoders: an audio-only model and a video-only model (see Fig.~\ref{fig:ablation_full}, left). 
This setup is equivalent to training separate A-JEPA~\cite{ajepa_2024} and V-JEPA models~\cite{vjepa_2024}, and serves as reference points to answer the central question: can learning from audio-visual synergy improve both modalities beyond what each achieves independently?
The following input representation and intra-modal prediction objective are used for these baselines and all subsequent experiments.

\paragraph{Input Representation.} Audio and video are tokenized by separate linear projection layers (a 2D convolution for audio spectrograms and a 3D convolution for video tubelets) that map each modality into the model's embedding space. 
To distinguish modalities and encode spatial or temporal structure, we add modality-specific embeddings and separate positional embeddings for each modality. 
Further details on tokenization, embeddings, and masking are provided in the supplementary material.

\paragraph{Intra-modal Prediction.} The primary training signal is the JEPA objective from Sec.~\ref{sec:preliminaries}, applied independently to each modality, which we refer to as the \textbf{intra-modal loss} (see Fig.~\ref{fig:architecture}a). 
For a given input mode $m \in \{a, v, av\}$, let $x^m_{\text{vis}}$ denote the visible (unmasked) tokens of modality $m$. 
The context encoder produces $z^m_{\text{vis}} = E_\theta(x^m_{\text{vis}})$, while the target encoder computes $y^m_{\text{mask}} = E_{\bar{\theta}}(x^m)_{\text{mask}}$ from the full, unmasked input. 
The predictor $P_\phi$ takes the encoded visible tokens along with learnable mask tokens $\Delta^m$ and predicts the target representations at the masked positions:
\begin{equation}
    \mathcal{L}_{m \to m} = \| P_\phi(z^m_{\text{vis}}, \Delta^m) - \text{sg}(y^m_{\text{mask}}) \|_1
\end{equation}
This yields three intra-modal loss terms: $\mathcal{L}_{a \to a}$, $\mathcal{L}_{v \to v}$, and $\mathcal{L}_{av \to av}$.
To encourage the learning of a rich feature hierarchy, we use multi-level prediction for the intra-modal objective, where features from multiple intermediate layers of the encoder are fused and predicted.
We provide implementation details on this multi-level prediction architecture in the supplementary material.

\paragraph{Training and Results.}
The \textbf{Audio-only Encoder} is trained on the audio of AS2M using only the audio intra-modal loss ($\mathcal{L}_{a \to a}$), achieving 30.87 mAP on audio evaluation. 
Similarly, the \textbf{Video-only Encoder} is trained on the video of AS2M using only the video intra-modal loss ($\mathcal{L}_{v \to v}$), achieving 19.84 mAP on video evaluation. 
The relatively lower video-only score is expected, as AS20K is an audio event classification benchmark where visual information alone may not be sufficient.

\subsection{Shared Encoder}
\label{sec:shared_encoder}
Since audio and video are different views of the same physical phenomena, their high-level representations should inhabit a common space~\cite{platonic_2024}. 
A natural architectural choice to encourage this is to use a \textbf{single, shared encoder} $E_\theta$ and \textbf{shared predictor} $P_\phi$ for both modalities, making parameter sharing an inductive bias toward a unified representation space. 

We test this hypothesis by training a single shared encoder-predictor pair on both audio and video from AS2M simultaneously. 
The model is trained with only the two intra-modal losses ($\mathcal{L}_{a \to a}$ and $\mathcal{L}_{v \to v}$), applied to their respective modalities. 
Audio and video are encoded independently and never jointly; there is no explicit cross-modal alignment objective. 

As shown in Fig.~\ref{fig:ablation_full} (shaded region), this naive approach is not effective. 
While the audio-video evaluation benefits from seeing both modalities at probe time (32.38 mAP), 
the individual modality performances \emph{drop below} their unimodal baselines: audio falls from 30.87 to 28.70 mAP and video from 19.84 to 17.90 mAP. 
Without an explicit alignment objective, the shared encoder weights are pulled in two directions --- one for each modality, leading to degraded representations for both modalities. 
This finding is consistent with prior work that also observed unimodal performance degradation when moving to a shared encoder~\cite{xkd_2022, lin2024avsiam, NEURIPS2021_VATT} and highlights the central challenge our work addresses.

\subsection{Shared Encoder $+$ Cross-modal Alignment}
\label{sec:cross_modal_alignment}
Recent work suggests that different modalities naturally converge towards a shared statistical model of reality~\cite{platonic_2024}, and that explicitly enforcing this alignment acts as a powerful learning signal that can synergize training across modalities~\cite{repa_2024, synergy_2025}. 
Motivated by this, we add two cross-modal prediction objectives ($\mathcal{L}_{a \to v}$, $\mathcal{L}_{v \to a}$) to the shared encoder from the previous step, which retains its intra-modal losses ($\mathcal{L}_{a \to a}$, $\mathcal{L}_{v \to v}$). 
The model is now trained with four losses in total, enforcing both modality-specific coherence and cross-modal semantic alignment. 
As illustrated in Fig.~\ref{fig:architecture}b, we realize this through a set of lightweight cross-modal predictors, $C_\psi^{m_1 \to m_2}$, which are simple MLPs that take the mean-pooled, last-layer output of the masked source modality to predict the mean-pooled, last-layer output of the unmasked target modality. 
The \textbf{cross-modal loss} for a source modality $m_1$ and target modality $m_2$ is:
\begin{equation}
    \mathcal{L}_{m_1 \to m_2} = \| C_\psi^{m_1 \to m_2}(\text{pool}(z^{m_1}_{\text{vis}})) - \text{sg}(\text{pool}(y^{m_2})) \|_1
\end{equation}
where $z^{m_1}_{\text{vis}} = E_\theta(x^{m_1}_{\text{vis}})$ is the context encoder's \emph{last-layer} output from the \emph{masked} source input, $y^{m_2} = E_{\bar{\theta}}(x^{m_2})$ is the target encoder's \emph{last-layer} output from the \emph{full, unmasked} target input, and $\text{pool}(\cdot)$ denotes mean pooling over tokens. 
In contrast to the multi-level intra-modal prediction, the cross-modal predictors operate only on last-layer features, since low-level features tend to be modality-specific (\eg, spectral patterns in audio, textures in video), whereas high-level features encode semantic concepts shared across modalities and are thus the appropriate target for cross-modal alignment. We use global pooling because there is no natural token-level correspondence between audio and video, making a pooled, holistic representation the appropriate level for cross-modal alignment.
Note that audio and video are still encoded independently in this configuration, \ie the model does not process joint audio-video inputs.

As shown in Fig.~\ref{fig:ablation_full}, this single addition of cross-modal alignment resolves the conflict observed in the previous step and substantially improves performance across all three settings: audio climbs to 33.52 mAP (surpassing the unimodal baseline by +2.65), video recovers to 19.58 mAP, and audio-video reaches 36.34 mAP. 
This confirms that cross-modal alignment is the critical ingredient that enables a shared encoder to benefit from multimodal data.

\subsection{Full \mjepa: Joint Audio-Video Encoding}
\label{sec:full_mjepa}
Building on the shared encoder with cross-modal alignment from the previous step, we now enable the encoder to also process concatenated audio-video tokens as a third input mode (see Fig.~\ref{fig:architecture}a), adding the joint intra-modal loss ($\mathcal{L}_{av \to av}$) and the remaining four cross-modal losses ($\mathcal{L}_{a \leftrightarrow av}$, $\mathcal{L}_{v \leftrightarrow av}$), for a total of six cross-modal losses: $\mathcal{L}_{a \leftrightarrow v}$, $\mathcal{L}_{a \leftrightarrow av}$, and $\mathcal{L}_{v \leftrightarrow av}$. 
This allows the encoder to learn from the full audio-visual context during pre-training, with all representations aligned through the cross-modal prediction objectives. As shown in Fig.~\ref{fig:ablation_full}, the result is another substantial boost across the board: audio reaches 38.89 mAP (+8.02 over the unimodal baseline), video reaches 25.38 mAP (+5.54), and audio-video reaches 42.90 mAP. The final training objective is the unweighted sum of all nine loss terms: three intra-modal ($\mathcal{L}_{a \to a}$, $\mathcal{L}_{v \to v}$, $\mathcal{L}_{av \to av}$) and six cross-modal ($\mathcal{L}_{a \leftrightarrow v}$, $\mathcal{L}_{a \leftrightarrow av}$, $\mathcal{L}_{v \leftrightarrow av}$). This is our base \textbf{\mjepa ViT-L} model.

\subsection{Full \mjepa $+$ Scaling}
\label{sec:scaling}
A key advantage of our unified architecture is its ability to efficiently leverage heterogeneous datasets. 
Because \mjepa uses a shared encoder that operates in audio-only, video-only, and audio-video modes, incorporating additional unimodal data requires no architectural changes—one simply feeds in video-only samples and computes $\mathcal{L}_{v \to v}$. 
This is naturally enabled by our shared encoder design, whereas prior AV-SSL methods that rely on separate encoders and paired data may require non-trivial modifications to incorporate unimodal sources. In practice, we train on both paired audio-visual and video-only data simultaneously by partitioning compute across the two data sources; implementation details are provided in the supplementary material.

Starting from the base ViT-L trained on AS2M (${\sim}$5K hours of paired audio-video data), we augment training with the video-only VM2M dataset~\cite{vjepa_2024}, adding ${\sim}$2M videos (${\sim}$136K hours) from standard video benchmarks (\textbf{+ Data Scaling}). As shown in Fig.~\ref{fig:ablation_full}, this significantly improves all metrics (mAP): audio from 38.89 to 40.00, video from 25.38 to 29.63, and audio-video from 42.90 to 45.31. 
Notably, video-only data improves not just video but also audio representations, confirming the positive cross-modal transfer enabled by our alignment objectives.
We then scale from ViT-L (300M) to ViT-g (1B parameters) (\textbf{+ Model Scaling}), yielding a final boost to 40.97 (A), 29.82 (V), and 45.44 (AV), demonstrating that \mjepa scales effectively with both data and model size.

%% file: sec/5_main_results.tex
\section{Main Results}
\label{sec:main_results}

We now evaluate our full \mjepa models against state-of-the-art self-supervised methods on a comprehensive set of audio, video, and audio-video classification benchmarks.
All evaluations use the frozen attentive probe protocol described in Sec.~\ref{sec:frozen_eval}, providing a fair and consistent comparison across methods.
The key takeaway from our results is that \mjepa's simple, unified architecture consistently learns superior frozen representations that outperform prior specialized models, and in several cases, even match or exceed fully finetuned models.
We first evaluate audio-video classification on AudioSet-20K (Sec.~\ref{sec:as20k_av_eval}), then audio-only performance on AS20K, ESC-50, and FSD50K (Sec.~\ref{sec:audio_frozen}), and finally video-only performance on Kinetics-400 and SSv2 (Sec.~\ref{sec:video_frozen}).
We additionally validate the quality of our cross-modal alignment through a retrieval experiment in the supplementary material, providing further evidence that our predictive objectives learns semantically meaningful cross-modal correspondences.

\subsection{Audio-video frozen evaluation on AudioSet20K.}
\label{sec:as20k_av_eval}
\input{tables/audioset20k_av_eval}

\paragraph{\textbf{Setting.}}
We evaluate on AudioSet-20K~\cite{audioset_2017} using the frozen evaluation protocol from Sec.~\ref{sec:frozen_eval}.
For baselines with separate audio and video encoders, we concatenate their output tokens before feeding them to the probe for the A-V setting.
To ensure a fair comparison, we evaluate all baselines with our attentive probe, which we found to be consistently stronger than linear probing (\eg, improving CAV-MAE Sync~\cite{cavmae_sync_cvpr25} from a reported 8.7 to 21.66 mAP).
Notably, all baselines use ImageNet-1K~\cite{russakovsky2015imagenetlargescalevisual} pretraining, whereas \mjepa is trained from scratch.

\paragraph{\textbf{Comparison with baselines.}} As shown in Table~\ref{tab:as20k_av_main}, our method sets a new state of the art for frozen audio-visual representations.
Our base \mjepa ViT-L model, trained only on AudioSet, already surpasses the previous best frozen model, EquiAV~\cite{equiav_icml24}, by a large margin: +4.6 mAP on audio-only, +6.8 on video-only, and +4.3 on audio-video evaluation.
Augmenting the pre-training with additional video data (AS+VM2M) further improves performance across all three modalities, demonstrating that the improved video representation from our unified encoder also enhances audio and audio-video features through cross-modal prediction objective.
This data-scaled \mjepa ViT-L model produces frozen features that are highly competitive with fully finetuned prior work.
Its frozen video-only performance (29.63 mAP) surpasses that of all reported finetuned baselines, including EquiAV (25.70 mAP).
Finally, by scaling the model size to ViT-g, \mjepa further closes the gap to the overall state of the art, achieving an audio-video mAP of 45.44, within 1.2 points of the fully finetuned EquiAV, while maintaining superior frozen video-only performance.
These results demonstrate that a simple JEPA-only objective can produce highly transferable representations that scale effectively with both data and model size.

\subsection{Audio-only frozen evaluation.}
\label{sec:audio_frozen}
\input{tables/audio_frozen}

\paragraph{\textbf{Setting.}}
We evaluate frozen audio representations on three tasks: multi-label sound event classification on AS20K~\cite{audioset_2017} and FSD50K~\cite{fsd50k_2022}, and single-label environmental sound classification on ESC-50~\cite{ESC50_2015}, using their standard splits.
Notably, FSD50K evaluation labels are considered clean due to an extensive multi-stage human validation process, making it a more reliable benchmark than AS20K.
Following Section~\ref{sec:frozen_eval}, we train an attentive probe over frozen features.
To ensure a fair comparison, we re-evaluated all publicly available baselines using our probe, which we found to be significantly stronger than the linear probes reported in many prior works, a finding also supported by recent analysis~\cite{unmute_2026} (\eg, improving SSLAM~\cite{sslam_2025} AS20K mAP from reported 16.9 to 31.4).
For models without public weights (MAViL~\cite{mavil_neurips23}, XKD~\cite{xkd_2022}) or where our probing yielded lower scores than reported (SPEAR~\cite{spear_iclr2026} on ESC-50 and FSD50K), we conservatively report their published numbers.

\paragraph{\textbf{Results and Analysis.}}
\label{sec:video_frozen}
As shown in Table~\ref{tab:audio_frozen}, \mjepa establishes a new state of the art for frozen representations on general audio tasks.
Our base ViT-L model, trained only on AudioSet, already outperforms all prior audio-only and audio-visual SSL methods on AS20K, ESC-50, and FSD50K.
This demonstrates that our unified audio-visual architecture learns superior audio features, due to the learned semantic correspondence between modalities by the intra-modal and cross-modal JEPA objectives.
This hypothesis is further supported by our data scaling results: augmenting the pre-training with additional video-only data (AS+VM2M) monotonically improves performance on all three audio tasks, demonstrating positive transfer from video to audio through our shared encoder and cross-modal alignment.
Notably, the frozen representations from our scaled models are competitive with, and in some cases surpass, fully finetuned models.
Our frozen \mjepa ViT-g achieves 96.9\% accuracy on ESC-50 and 65.8 mAP on FSD50K, outperforming the best finetuned results reported from XKD~\cite{xkd_2022} (96.5\%) and EquiAV~\cite{equiav_icml24} (62.6\%), respectively.

\subsection{Video-only frozen evaluation.}
\input{tables/video_frozen}

\paragraph{\textbf{Setting.}}
We evaluate the quality of the learned frozen video representations on two standard video understanding benchmarks: Kinetics-400 (K400)~\cite{kinetics_2017}, which focuses on appearance understanding, and Something-Something-v2 (SSv2)~\cite{ssv2_2017}, a more challenging benchmark for motion understanding.
Following the protocol of VJEPA~\cite{vjepa_2024} and VJEPA2~\cite{vjepa2_2025}, we report top-1 accuracy using an attentive probe trained on frozen features with 16 frames, $8$ segments ${\times}$ $3$ spatial views for K400 and $2$ segments ${\times}$ $3$ spatial views for SSv2.
Our comparison focuses on state-of-the-art video-only SSL methods, as prior AV-SSL methods are typically trained only on AudioSet and not on video-centric datasets.

\paragraph{\textbf{Results and Analysis.}}
Table~\ref{tab:video_frozen} highlights the benefit of learning from aligned audio-visual data.
We first note that \mjepa trained solely on AudioSet~\cite{audioset_2017}, without any video-specific data, already achieves reasonable video recognition performance (75.2\% on K400, 58.6\% on SSv2), indicating that the audio-visual correspondence in AudioSet provides a useful learning signal for video understanding.
When trained on video-only VM2M data using only the $\mathcal{L}_{v \to v}$ objective, our ViT-L model achieves 80.6\% on K400 and 69.8\% on SSv2, on par with the dedicated video-only VJEPA model (80.8\% / 69.5\%), demonstrating that our unified architecture does not compromise video representation quality.
The true advantage of our approach becomes evident when we combine both data sources.
By training on AS+VM2M, the K400 performance of our ViT-L model jumps by over 4 points to 84.7\%, significantly surpassing the VM2M-only baseline.
This improvement stems from the cross-modal JEPA objective, which aligns representations between modalities, acting as a complementary learning signal that enriches the learned video representations.
Consequently, our ViT-L model trained on AS+VM2M nearly matches the performance of VJEPA2 ViT-L on both K400 (84.7\% vs.\ 85.1\%) and SSv2 (73.3\% vs.\ 73.7\%), despite VJEPA2 being pre-trained on VM22M~\cite{vjepa2_2025}, a dataset with approximately $10\times$ more video data.
Scaling to ViT-g further improves performance to 85.0\% on K400 and 73.9\% on SSv2, underscoring the scalability of our approach.

%% file: tables/audioset20k_av_eval.tex
\begin{table}[t]
\centering
\caption{\textbf{AudioSet-20K audio-visual evaluation.} We report mAP ($\uparrow$) for audio-only (A), video-only (V), and audio-video (A-V) inputs. All frozen results use the same attentive probing protocol; finetuned results are from respective papers. \textbf{Bold}: best frozen; \uline{underline}: best overall. \mjepa achieves frozen state of the art across all three settings, with frozen video-only results surpassing all finetuned baselines.}
\vspace{-3mm}
\label{tab:as20k_av_main}
\begin{adjustbox}{max width=\textwidth}
\begin{threeparttable}
\begin{tabular}{lcc|ccc}
\toprule
& & & \multicolumn{3}{c}{AudioSet20K mAP $\uparrow$} \\
\cmidrule(lr){4-6}
Method & Params & Pre-train data & A & V & A-V \\
\midrule
\textit{Frozen evaluation} & & & & & \\
\addlinespace[2pt]
CAV-MAE~\cite{cavmae_iclr23} & 170M & IN+AS & 19.38 & 18.14 & 34.59 \\
MAViL~\cite{mavil_neurips23}\tnote{a} & 170M & IN+AS & 30.00 & -- & -- \\
EquiAV~\cite{equiav_icml24} & 170M & IN+AS & 34.25 & 18.60 & 38.60 \\
CAV-MAE Sync~\cite{cavmae_sync_cvpr25} & 170M & IN+AS & 21.66 & 16.20 & 28.50 \\
\addlinespace[2pt]
\rowcolor{ourgray}
MJEPA ViT-L (ours) & 300M & AS & 38.89 & 25.38 & 42.90 \\
\rowcolor{ourgray}
MJEPA ViT-L + data scaling (ours) & 300M & AS+VM2M & 40.00 & 29.63 & 45.31 \\
\rowcolor{ourgray}
MJEPA ViT-g + data scaling (ours) & 1B & AS+VM2M & \textbf{40.97} & \uline{\textbf{29.82}} & \textbf{45.44} \\
\midrule
\textit{Full-model finetuning (reported)} & & & & & \\
\addlinespace[2pt]
CAV-MAE~\cite{cavmae_iclr23} & 170M & IN+AS & 37.70 & 19.80 & 42.00 \\
MAViL~\cite{mavil_neurips23} & 170M & IN+AS & 41.80 & 24.80 & 44.90 \\
EquiAV~\cite{equiav_icml24} & 170M & IN+AS & \uline{42.40} & 25.70 & \uline{46.60} \\
\bottomrule
\end{tabular}
\begin{tablenotes}[flushleft]
\footnotesize
\item[a] \scriptsize MAViL does not release pretrained weights; the paper only reports frozen audio-only result.
\end{tablenotes}
\end{threeparttable}
\end{adjustbox}
\vspace{-3mm}
\end{table}

%% file: tables/audio_frozen.tex
\begin{table}[t]
\centering
\caption{\textbf{Audio-only frozen evaluation.} We report AS20K and FSD50K mAP ($\uparrow$), and ESC-50 accuracy ($\uparrow$). \textbf{Bold}: best frozen result; \uline{underline}: best overall. \mjepa sets a new state of the art for frozen audio representations, even surpassing several fully finetuned methods.}
\vspace{-3mm}
\label{tab:audio_frozen}
\begin{adjustbox}{max width=\textwidth}
\begin{threeparttable}
\begin{tabular}{lclc|ccc}
\toprule
Method & Encoder Params & Modality & Pre-train data
& AS20K & ESC-50 & FSD50K \\
\midrule

\multicolumn{7}{l}{\textbf{Frozen feature probing}} \\
\midrule
AJEPA~\cite{ajepa_2024} & 85M & A & AS & 18.0 & 74.4 & 43.9 \\
SSLAM~\cite{sslam_2025} & 88M & A & AS & 31.4 & 93.2 & 57.9 \\
SPEAR~\cite{spear_iclr2026}\tnote{*} & 600M & A & 197k hrs mix & 11.9 & 89.4 & 57.1 \\
Dasheng-1.2B~\cite{dasheng_interspeech2024}\tnote{*} & 1.2B & A & 272k hrs mix & 31.6 & 92.2 & 56.9 \\
\midrule
CAV-MAE~\cite{cavmae_iclr23} & 170M & A+V & IN+AS & 19.4 & 77.5 & 46.1 \\
MAViL~\cite{mavil_neurips23} & 170M & A+V & IN+AS & 30.0 & 90.8 & -- \\
XKD~\cite{xkd_2022} & 170M & A+V & AS & -- & 93.6 & 51.5 \\
EquiAV~\cite{equiav_icml24} & 170M & A+V & AS & 34.3 & 93.2 & 57.9 \\
CAV-MAE Sync~\cite{cavmae_sync_cvpr25} & 170M & A+V & AS & 21.7 & 89.2 & 55.5 \\
\midrule
\rowcolor{ourgray}
MJEPA ViT-L (ours) & 300M & A+V & AS & 38.9 & 95.2 & 63.9 \\
\rowcolor{ourgray}
MJEPA ViT-L + data scaling (ours) & 300M & A+V & AS+VM2M & 40.0 & 96.8 & 65.5 \\
\rowcolor{ourgray}
MJEPA ViT-g + data scaling (ours) & 1B & A+V & AS+VM2M & \textbf{40.9} & \textbf{\uline{96.9}} & \textbf{\uline{65.8}} \\

\midrule
\multicolumn{7}{l}{\textbf{Full-model finetuning (reported)}} \\
\midrule
AJEPA~\cite{ajepa_2024} & 85M & A & AS & 38.4 & 96.3 & -- \\
SSLAM~\cite{sslam_2025} & 88M & A & AS & 40.9 & 96.2 & -- \\
SPEAR~\cite{spear_iclr2026}\tnote{*} & 600M & A & 197k hrs mix & 39.4 & -- & -- \\
CAV-MAE~\cite{cavmae_iclr23} & 170M & A+V & IN+AS & 37.7 & -- & -- \\
MAViL~\cite{mavil_neurips23} & 170M & A+V & IN+AS & 41.8 & 94.4 & -- \\
XKD~\cite{xkd_2022} & 170M & A+V & AS & -- & 96.5 & 58.5 \\
EquiAV~\cite{equiav_icml24} & 170M & A+V & AS & \uline{42.4} & 96.0 & 62.6 \\
\bottomrule
\end{tabular}
\begin{tablenotes}[flushleft]
\footnotesize
\item[*] \scriptsize SPEAR~\cite{spear_iclr2026} and Dasheng~\cite{dasheng_interspeech2024} use significantly more pre-training data (197k and 272k hours respectively) compared to the 5k hours of audio in AudioSet. SPEAR also uses Dasheng~\cite{dasheng_interspeech2024} and WavLM~\cite{WavLM_2022} as teachers.
\end{tablenotes}
\end{threeparttable}
\end{adjustbox}
\vspace{-2mm}
\end{table}

%% file: tables/video_frozen.tex
\begin{table}[t]
\centering
\caption{\textbf{Video-only frozen evaluation.} We report top-1 accuracy ($\uparrow$) on Kinetics-400 (appearance-focused) and Something-Something v2 (SSv2; motion-focused). \textbf{Bold}: best result per column. Despite using $10\times$ less video data than VJEPA2, \mjepa achieves comparable performance by leveraging cross-modal alignment with audio.}
\vspace{-3mm}
\label{tab:video_frozen}
\begin{adjustbox}{max width=\columnwidth}
\begin{threeparttable}
\begin{tabular}{lcc|cc}
\toprule
\cmidrule(lr){4-5}
Method & Encoder Params & Pre-train data & K400 & SSv2 \\
\midrule
VJEPA ViT-L~\cite{vjepa_2024} & 300M & VM2M & 80.8 & 69.5 \\
VJEPA ViT-H~\cite{vjepa_2024} & 600M & VM2M & 82.0 & 71.4 \\
VJEPA2 ViT-L~\cite{vjepa2_2025} & 300M & VM22M & 85.1 & 73.7 \\
VJEPA2 ViT-g~\cite{vjepa2_2025} & 1B & VM22M & \textbf{86.6} & \textbf{75.3} \\
\midrule
\rowcolor{ourgray}
MJEPA ViT-L (ours) & 300M & AS & 75.2 & 58.6 \\
\rowcolor{ourgray}
MJEPA ViT-L (ours)\tnote{a} & 300M & VM2M & 80.6 & 69.8 \\
\rowcolor{ourgray}
MJEPA ViT-L + data scaling (ours) & 300M & AS+VM2M & 84.7 & 73.3 \\
\rowcolor{ourgray}
MJEPA ViT-g + data scaling (ours) & 1B & AS+VM2M & 85.0 & 73.9 \\
\bottomrule
\end{tabular}
\begin{tablenotes}[flushleft]
\footnotesize
\item[a] \scriptsize Trained with video-only intra-modal loss ($\mathcal{L}_{v \to v}$) for comparison with VJEPA.
\end{tablenotes}
\end{threeparttable}
\end{adjustbox}
\vspace{-2mm}
\end{table}

%% file: sec/6_conclusion.tex
\section{Conclusion}
\label{sec:conclusion}
In this work, we introduced \mjepa, a simple and scalable framework for audio-visual self-supervised learning. 
Our approach departs from the prevailing trend of complex, multi-component architectures by demonstrating that a single, unified encoder trained with only a JEPA objective is sufficient to learn powerful representations. 
We showed that while naively sharing an encoder degrades performance, the addition of a cross-modal prediction objective is the critical ingredient that enables positive transfer, allowing each modality to synergistically improve the other and enabling the framework to scale effectively with both model size and heterogeneous data. 
The resulting frozen representations set a new state of the art for audio-visual SSL, and their strong performance against fully finetuned models underscores their generality and robustness, a key advantage over more task-specific, adapted representations.

Our work opens promising avenues for future research. 
The surprising effectiveness of our simple, global-pooled cross-modal predictors suggests that more expressive cross-modal prediction mechanisms could learn even richer alignment across modalities.
While this work focuses on audio and video, \mjepa serves as a successful proof-of-concept for a truly modality-agnostic architecture, paving the way for unified models in domains like medical imaging (\eg, CT, MRI, X-ray) or robotics (\eg, vision, proprioception) that can flexibly learn from co-occurring data and are robust to missing modalities at test time.

%% file: sec/7_appendix.tex
\section*{Supplementary Material}
This supplementary material provides model and implementation details (Sec.~\ref{sec:model_details}), training details (Sec.~\ref{sec:training_details}), and an additional cross-modal retrieval evaluation (Sec.~\ref{sec:retrieval_eval}).

\section{Model and Implementation Details}
\label{sec:model_details}

\begin{figure*}[t]
    \centering
    \includegraphics[width=0.85\textwidth]{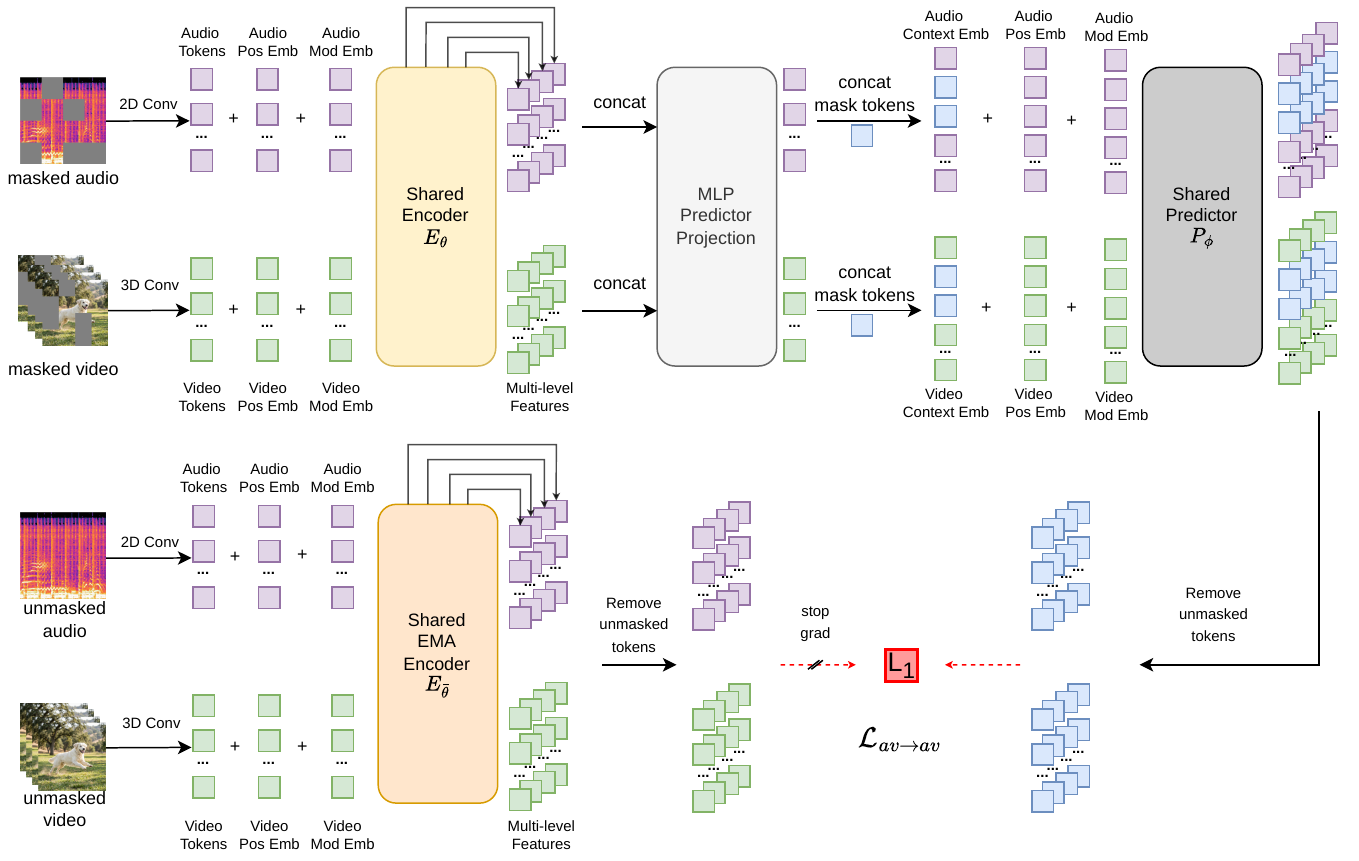}
    \caption{\textbf{Detailed architecture of \mjepa}, illustrated for the joint audio-video intra-modal prediction ($\mathcal{L}_{av \to av}$). The audio-only ($\mathcal{L}_{a \to a}$) and video-only ($\mathcal{L}_{v \to v}$) cases follow the same pipeline with a single modality branch. \textbf{Top-left:} Masked audio and video inputs are tokenized by modality-specific projection layers (2D Conv for audio, 3D Conv for video), and augmented with modality-specific positional and modality embeddings before being fed to the shared context encoder $E_\theta$. \textbf{Top-middle:} Multi-level features from intermediate encoder layers are concatenated along the embedding dimension and projected back to the model dimension via a MLP. \textbf{Top-right:} The projected context tokens are concatenated with learnable mask tokens, added with predictor-specific modality embeddings and positional embeddings, and processed by the shared predictor $P_\phi$ to produce multi-level predictions for the masked positions. \textbf{Bottom:} The target branch processes the full, unmasked inputs through the EMA encoder $E_{\bar{\theta}}$ and extracts multi-level features at the masked positions. The $L_1$ loss is computed between the predicted and target representations with a stop-gradient on the target. Cross-modal predictors are not shown; they operate on pooled last-layer features as described in Sec.~\ref{sec:cross_modal_alignment} of the main paper.}
    \label{fig:detailed_architecture}
\end{figure*}

Fig.~\ref{fig:detailed_architecture} provides a detailed view of the \mjepa architecture, complementing the high-level overview in Fig.~\ref{fig:architecture} of the main paper. We describe the components in detail below.

\paragraph{Encoder Architecture.} The \mjepa encoder follows the architecture of the vision transformer~\cite{dosovitskiy2021image}, and we propose two sizes: ViT-L (300M) and ViT-g (1B).

\paragraph{Input Tokenization and Embeddings.} As shown in Fig.~\ref{fig:detailed_architecture}, inputs are tokenized by modality-specific layers—a 2D convolution for audio spectrograms and a 3D convolution for video tubelets—that project patches or tubelets into the shared latent embedding space. 
To inform the shared encoder of the input structure, we add modality-specific learnable embeddings and positional embeddings. 
We use 2D sincos absolute positional embeddings for audio tokens (time, frequency) and 3D sincos for video tokens (time, height, width).

\paragraph{Predictor Architectures.} The shared intra-modal predictor is a narrow ViT (384 embedding dimension) that receives multi-level features from the encoder. 
Intermediate features from layers \{5, 11, 17, 23\} for ViT-L, and \{9, 19, 29, 39\} for ViT-g, are concatenated along the embedding dimension and fused via a 2-layer MLP before being passed to the predictor. 
The predictor then adds its own set of modality-specific positional and modality embeddings before processing the sequence. 
These positional embedding are same as encoder positional embedding - 2D sincos absolute positional embeddings for audio tokens and 3D sincos for video tokens.
In contrast, the six cross-modal predictors are simple 3-layer MLPs that operate only on the mean-pooled, last-layer features of the encoder.

\paragraph{Attentive Probe Architecture.} Our attentive probe consists of a learnable query token that attends to all encoder output tokens via cross-attention, followed by 4 transformer blocks and a linear classification head.

\section{Training Details}
\label{sec:training_details}

\begin{table*}[t]
\centering
\caption{\textbf{Key training hyperparameters for MJEPA models.} AS2M denotes the audio-visual AudioSet-2M dataset, while VM2M denotes the video-only VideoMix2M dataset. The scaled models are pre-trained with a constant learning rate and then undergo a cooldown phase with an increased frame count and a linear learning rate decay.}
\resizebox{\textwidth}{!}{
\begin{tabular}{@{}l|ccc@{}}
\toprule
\textbf{Hyperparameter} & \textbf{Base ViT-L} & \textbf{Scaled ViT-L} & \textbf{Scaled ViT-g} \\
\midrule
Datasets & AS2M & AS2M + VM2M & AS2M + VM2M \\
Total Batch Size & 512 & 2560 (512 AV + 2048 V) & 2880 (576 AV + 2304 V) \\
Crop Size & 224 & 256 & 256 \\
LR Schedule & Cosine decay (6e-4 to 1e-6) & Constant (5.25e-4) & Constant (5.25e-4) \\
EMA Schedule & Cosine decay (0.999 to 1.0) & Constant (0.999) & Constant (0.999) \\
Pre-training Steps & 150k & 150k & 150k \\
Cooldown Steps & - & 12k & 12k \\
Cooldown Frames & - & 64 & 64 \\
\bottomrule
\end{tabular}
}
\label{tab:hyperparams}
\end{table*}

\paragraph{Optimizer and Precision.} All models are trained with the AdamW optimizer~\cite{adamw_2019} with $\beta_1=0.9$ and $\beta_2=0.995$. We use bfloat16 precision for all training and evaluation.

\paragraph{Preprocessing and Augmentation.} We follow standard data preprocessing practices. For audio, we adopt the pipeline from AST~\cite{AST_interspeech2021}, converting each 10-second clip into a $1024 \times 128$ log-mel spectrogram, which is then normalized using a mean of -4.268 and a standard deviation of 4.569. 
For video, we use random resized cropping with a scale of [0.3, 1.0] and an aspect ratio of [0.75, 1.35]. 
The crop size is 224 for the base ViT-L model and 256 for the scaled models.

\paragraph{Masking Strategy.} We use modality-specific masking. 
For audio, we randomly mask 75-80\% of patches. For video, our base ViT-L model uses a single masking strategy of 8 small blocks (15\% of the frame) that span the full temporal dimension. 
For the scaled models (ViT-L and ViT-g), we adopt the more aggressive multi-block strategy from VJEPA2~\cite{vjepa2_2025}, applying a combination of eight smaller masks (15\% of the frame) and two larger masks (70\% of the frame) per iteration. 
For combined audio-video inputs, masking is applied independently to each stream.

\paragraph{Training Schedule.} The base ViT-L model is trained on AS2M for 150k steps with a batch size of 512. 
It uses a cosine learning rate decay from 6e-4 to 1e-6 with a 60-epoch warmup, and a cosine EMA schedule from 0.999 to 1.0. 
The scaled models (ViT-L and ViT-g on AS2M+VM2M) are pre-trained for 150k steps with a constant learning rate of 5.25e-4 and a fixed EMA of 0.999, both with a 40-epoch warmup. 
This is followed by a 12k-step cooldown phase where the input is increased to 64 frames and the learning rate is linearly decayed to zero.

\paragraph{Distributed Training for Data Scaling.} When training with additional video-only data, we partition the workload across GPUs: a subset of GPUs processes paired audio-visual samples, while the remaining GPUs process video-only samples, computing only $\mathcal{L}_{v \to v}$. 
Gradients from all GPUs are combined via all-reduce before a single optimizer step. 
The sampling weights for the video-only datasets (Kinetics-710, SSv2, HowTo100M) follow those used in V-JEPA~\cite{vjepa_2024}. 
A scaling factor of 5.0, determined via hyperparameter search, is applied to the video-only loss to balance gradient magnitudes across the two data sources.

\section{Cross-modal Retrieval Evaluation}
\label{sec:retrieval_eval}

To further validate the quality of the cross-modal alignment learned by \mjepa, we evaluate the cross-modal predictors on a retrieval task.
Unlike prior methods that rely on explicit contrastive objectives designed for retrieval, \mjepa is trained with a pure predictive objective.

Given a set of audio-video pairs, we first extract audio and video features independently using our shared \mjepa encoder — audio tokens are fed to obtain audio features, and video tokens are fed to obtain video features.
For audio-to-video retrieval, we pass the audio features through the cross-modal predictor $C_\psi^{a \to v}$ to generate a predicted video representation, and retrieve the nearest ground-truth video representation from the pool using L1 distance, consistent with our training objective.
The same procedure applies for video-to-audio retrieval using $C_\psi^{v \to a}$.
We compute distances on pre-final layer norm features.
Baselines typically use cosine similarity, which is natural for their contrastive losses.
Following prior work~\cite{cavmae_iclr23, cavmae_sync_cvpr25, equiav_icml24}, we evaluate on the AudioSet retrieval subset.

\input{tables/retrieval}

As shown in Table~\ref{tab:retrieval}, \mjepa, without any contrastive objective, achieves retrieval performance that is highly competitive with state-of-the-art contrastive methods such as EquiAV~\cite{equiav_icml24}.
This is a notable finding, as prior work has shown that removing the contrastive objective can cause retrieval performance to collapse to random chance~\cite{cavmae_iclr23}.
While methods like CAV-MAE Sync~\cite{cavmae_sync_cvpr25}, which use explicit temporal alignment and a contrastive loss, show stronger video-to-audio performance, our results demonstrate that the JEPA predictive objective alone is sufficient to learn strong semantic alignment between modalities.

%% file: tables/retrieval.tex
\begin{table}[t]
\centering
\caption{\textbf{Cross-modal retrieval on AudioSet.} We report recall at $K$ (R@1/5/10; $\uparrow$) for audio$\rightarrow$video and video$\rightarrow$audio retrieval. Despite using no contrastive objective, \mjepa achieves retrieval performance competitive with state-of-the-art contrastive methods, validating the quality of our cross-modal alignment.}
\label{tab:retrieval}
\begin{adjustbox}{max width=\columnwidth}
\begin{threeparttable}
\begin{tabular}{l|ccc|ccc}
\toprule
& \multicolumn{3}{c|}{Audio$\rightarrow$Video R@K $\uparrow$} & \multicolumn{3}{c}{Video$\rightarrow$Audio R@K $\uparrow$} \\
\cmidrule(lr){2-4}\cmidrule(lr){5-7}
Method & R@1 & R@5 & R@10 & R@1 & R@5 & R@10 \\
\midrule
CAV-MAE~\cite{cavmae_iclr23} & 15.1 & 34.0 & 43.0 & 18.8 & 39.5 & 50.1 \\
CAV-MAE Sync~\cite{cavmae_sync_cvpr25} & 27.9 & 52.4 & 62.2 & \textbf{35.2} & \textbf{58.3} & \textbf{67.6} \\
EquiAV~\cite{equiav_icml24} & \textbf{29.6} & \textbf{53.7} & \textbf{63.1} & 30.1 & 53.3 & 62.9 \\
\rowcolor{ourgray}
MJEPA (ours) & 29.3 & \textbf{53.7} & 61.8 & 30.6 & 53.4 & 61.3 \\
\bottomrule
\end{tabular}
\end{threeparttable}
\end{adjustbox}
\end{table}